# Leveraging GenAI for Segmenting and Labeling Centuries-old Technical Documents

Carlos Monroy
*Department of Mathematics and Computer Science*
*University of St. Thomas*
Houston, Texas
monroyc@stthom.edu

Benjamin Navarro
*Department of Mathematics and Computer Science*
*University of St. Thomas*
Houston, Texas
navarrb@stthom.edu

*Abstract*—Image segmentation and image recognition are well established computational techniques in the broader discipline of image processing. Segmentation allows to locate areas in an image, while recognition identifies specific objects within an image. These techniques have shown remarkable accuracy with modern images, mainly because the amount of training data is vast. Achieving similar accuracy in digitized images of centuries-old documents is more challenging. This difficulty is due to two main reasons: first, the lack of sufficient training data, and second, because the degree of specialization in a given domain. Despite these limitations, the ability to segment and recognize objects in these collections is important for automating the curation, cataloging, and dissemination of knowledge, making the contents of priceless collections accessible to scholars and the general public. In this paper, we report on our ongoing work in segmenting and labeling images pertaining to shipbuilding treatises from the XVI and XVII centuries, a historical period known as the Age of Exploration. To this end, we leverage SAM2 [1] for image segmentation; Florence2 [2] and ChatGPT [3] for labeling; and a specialized ontology *ontoShip* and glossary *glosShip* of nautical architecture [4, 5] for enhancing the labeling process. Preliminary results demonstrate the potential of marrying these technologies for improving curation and retrieval of priceless historical documents. We also discuss the challenges and limitations encountered in this approach and ideas on how to overcome them in the future.

## I. INTRODUCTION

Digital collections in different domains have grown since the inception of the Internet. The availability of photographs has increased exponentially with the use of social media and mobile devices. This explains technology companies such as Meta, Google, Amazon, or Microsoft deploying sophisticated systems for indexing and retrieving images. At a fundamental level, images need to be associated with labels in order to identify what objects they depict and more broadly properly describe the scene in which they appear. With billions of images at their disposal, it is not surprising that these companies have advanced AI-powered image recognition to a new level. Take for example Meta's (Facebook parent company) SAM and SAM2 algorithms that can be used for segmenting images and videos. Blue edges in Fig. 1 correspond to the contours of objects detected by the SAM2 algorithm, noticeable are: table, chair, frames, and a plant to name a few. Results from this algorithm can be used to label objects or describe more complex scenes. We argue that these methods applied to modern photographs achieve impressive results; which is possible thanks to the variety, scope, and scale of the images used in training these systems. However, when applied to more specialized domains, the outcomes could be less accurate; this could be attributed to the lack of images for training as well as representing complex illustrations that are not easy to interpret, unless one has deep technical knowledge of that domain. Let us imagine a photograph in which the following objects are present: a person, a cat, a car, and a table. Meta's SAM2 will easily find the segments where those objects appear, helping to identify them, and facilitating an accurate description of the scene. A reasonable description could be: "person with pet eating at an outdoor restaurant," although other interpretations are also possible. In other disciplines, achieving such degree of detail might be more difficult because of the specialized knowledge required. In medicine, recognizing tumors and the degree of metastasis is possible because of the radiologist and oncologist expertise and the fact that these systems have been trained taking into consideration specific properties of images in oncology and radiology. Indeed, the use of machine learning to image recognition in cancer and medicine has advanced substantially in recent years.

In this paper we examine the application of image segmentation and labeling to a collection if images in a technical domain that poses interesting challenges. We propose an approach for improving labeling by leveraging terms in definitions from a specialized glossary along with knowledge expressed in a well-curated ontology. Our hypothesis is that this approach can be useful for other domains such as literature, art, music, and history; facilitating not only the retrieval of these materials, but also better understand their depictions. This paper is organized as follows: the importance of curation and dissemination of cultural heritage collections is highlighted in section II. Section III introduces nautical architecture and shipbuilding treatises. Image segmentation and labeling methodologies are described in Section IV. The use of the ontology and glossary is explained in section V. Results and ideas for future work are discussed in sections VI and VII respectively.

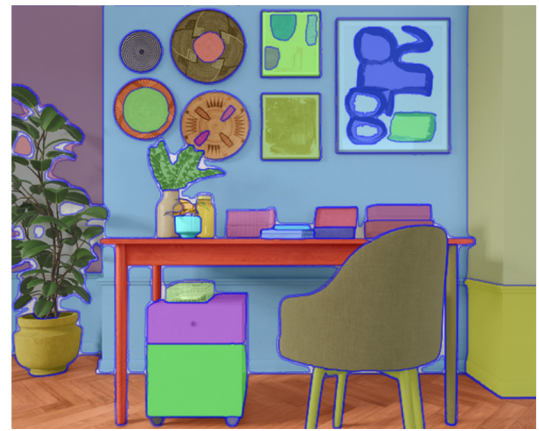

Fig. 1. Results of SAM2 segmentation algorithm depicting the contours of the objects found in a photograph (source SAM2 website).

II. DISSEMINATION OF CULTURAL HERITAGE COLLECTIONS

Images play a critical role in documenting daily life and have been present for thousands of years. From painting and engravings in caves, images were then printed on paper, or depicted in paintings and murals. In the twentieth century photography enabled to document scenes of daily life. More recently, the use of social media has produced quantities and diversity of images that were previously hard to imagine, paving the way for advancements in machine learning and image recognition. These images document specific historical or contemporary events and cultural periods; hence the value they have.

The possibility of indexing and searching in repositories of images is important not only for scholars interested in researching a particular topic, but also to the public for understanding historic events. For the past several decades, numerous efforts have been made to create digital collections of scholarly artifacts. Fr. Roberto Bussa, an Italian priest, persuaded IBM executives in the 1950's to allow him access to computing technology for the creation of a searchable archive of the works of Thomas Aquinas (The Index Thomisticus). Although it took more than three decades to be completed, Fr. Busa's work offers a great example of the way in which technology can be used to disseminate priceless collections. Arguably, Fr. Bussa is considered the precursor of the digital humanities discipline. Indeed, under the digital humanities umbrella, research and dissemination of numerous collections and repositories have been possible pertaining to both written [6] and graphical [7, 8, 9, 10, 11, 12, 13, 14, 15] artifacts.

III. NAUTICAL ARCHITECTURE

Naval architecture is the discipline that studies the construction of ships and the technical aspects related to their design. For millennia, ships were extremely important for commerce and trade. For centuries ships were the only means to transport goods and people across continents. Notably, during the age of exploration electronic instrumentation such as GPS, satellites, and radio were non-existent; therefore, navigation relied solely on captains' and sailors' expertise and skills coupled with limited knowledge about currents and winds. The construction techniques and structural knowledge developed during these period offer a wealth of information about these incredible achievements and the impact they had in exploring new territories, expanding commerce, and influencing different civilizations and cultures.

Because seafaring was an extremely dangerous enterprise, numerous ships were lost at sea. In this context, nautical archaeology studies ships and the people and cultures that created and used them. Archaeologists' work involves recovering ships pieces and components—most of the time damaged—due to the environment's impact and time elapsed since they sank. They also develop proper techniques to ensure preservation of ship remains that can lead to the reconstruction of sunken vessels.

The reconstruction process is challenging because parts are incomplete and damaged [4, 5, 16, 17]. Also, provenance of the ship is often unknown, and access to blueprints is nearly impossible, due to the difficulty to identify if a specific manual was used in its construction. Therefore, the main sources used to understand the ship is by comparing documentation from other excavations and existing documents.

A. Sources for Naval Architecture

Among the sources used in naval architecture are shipbuilding treatises. These were documents initially handwritten and later printed books. Although their content varies, for the most part, they include descriptions of the materials and parts required, the way in which to assemble them, and the sequences to build a ship, ensuring that it can sustain the winds, currents, crew, and cargo. Most treatises also include illustrations. These graphical representations play an important role since scholars can better understand construction techniques and sequences. Fig. 2. depicts illustrations in four different treatises written in diverse languages spanning approximately two centuries.

IV. IMAGE SEGMENTATION AND LABELING METHODOLOGIES

A. Identifying Segments in Images

Segmentation algorithms have been extensively used for image processing and image recognition. OpenCV [18] a well-known image processing library offers several implementations. The watershed algorithm [19, 20, 21] is a popular segmentation method based on ideas from topography to identify different segments in images. Segmenting images is a critical step in our pipeline because it enables to break down a composite image, identifying sections that correspond to specific objects. Meta recently released a segmentation algorithm called SAM2 (Segment Anything Model 2) [1], which has demonstrated great results. Fig. 3 shows segments from an illustration in a XVII century Portuguese treatise identified by SAM2. The identified segments can be cut out from the original image for improving indexing and further analysis.

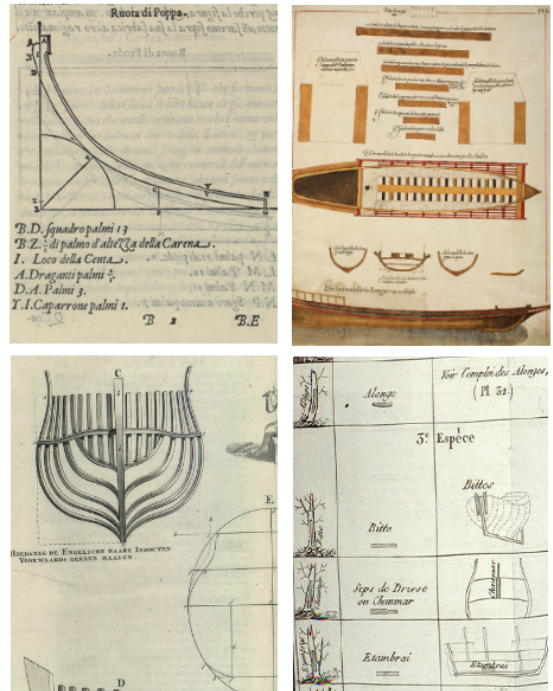

Fig. 2. Illustrations from four different treatises. From top to bottom in clockwise order Italian 1601, Portuguese 1616, French 1813, Dutch 1691.

The degree of granularity of the segments found in an image are important. For example, imagine a photograph of a street where there are trees, houses, people and a car. A high level of granularity would be identifying the car as an object in such image. However, since a car is a composite object, make out of different parts; identifying finer segments in the car, such as tires, windows, doors, and windshield offers more detailed information. The illustration in Fig. 4 corresponds to a Portuguese treatise depicting a composite object of a ship. A segmentation algorithm that identifies the five segments (Alifez, Quilha, Codaste, Escarva, and Patinha), performs better than one that identifies only the entire object (Couce da Popa). The former can be better indexed and retrieved.

*B. Proposed Segmentation and Labeling Pipeline*

Our methodology is based on a three-step pipeline. First, an algorithm is used to identify segments in the image. In the second step, labels are generated for the objects in the image. In the third step, bounding boxes are assigned to the different sections, and the labels are assigned to the appropriate bounding boxes. We have experimented with several approaches for step two, as described in the following paragraphs.

**Method 1:** BLIP created a caption for the image. NLTK was then used to generate, lemmatize, and separate one-word tags based on the caption for each object in the image. These tags were then fed to GroundingDINO [22], which created bounding boxes for each object detected, then assigned the BLIP/NLTK-generated tags to each bounding box. After this step, SAM was used to generate segments for each object in the tagged image.

**Method 2:** BLIP and NLTK were replaced with Tag2Text for generating image captions and tags, but this turned out not to be an ideal approach, because Tag2Text worked by using an embedding with a pre-included plaintext list of terms.

**Method 3:** We then used an earlier version of the Recognize Anything Model (RAM) to replace BLIP and NLTK. This is essentially an improved version of Tag2Text that also used a plaintext list of terms, but with an improved vocabulary and accuracy compared to Tag2Text. This approach uses SAM for segmentation and RAM for returning a list of labels for the image. Labels were then fed to GroundingDINO, generating bounding boxes on the image and assigning labels to each box.

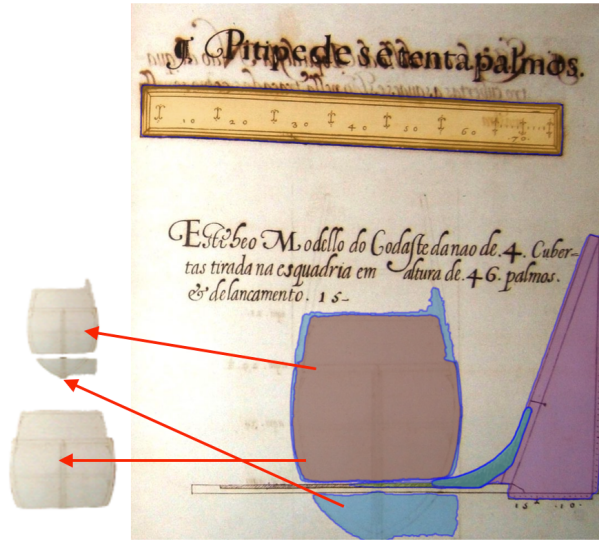

Fig. 3. Results of SAM2 segmentation applied to a 1616 Portuguese treatise. This tool not only detects the segments in the image shown in color, but it is also possible to extract the carved-out objects.

**Method 4:** Uses SAM to segment images, RAM to tag images, and GroundingDINO to generate boxes. Using openset inference, RAM builds an embedding with the detailed definitions generated by ChatGPT for each term in our glossary. It then generates the labels for the image with help from that embedding. The labels returned are then fed to GroundingDINO, which creates bounding boxes and assigns RAM's tags to them. Alternatively, the ChatGPT descriptions corresponding to RAM's tags could be fed to GroundingDINO as very long tags, instead of feeding the tags themselves. However, this produced much poorer and random results.

Segmentation results varied depending on the method use. Fig. 5 shows segments identified by SAM2 (left), note that the color offers more detailed information about the properties of such segments. Similarly, the heptagons inside the circular areas as well as two rectangles corresponding to bleed through images, and a text segment were also detected. Results using the Watershed algorithm (right) show a fewer number of areas identified. Similarly, only the exterior part of the circumferences was marked (red circle). Notably, the heptagons and text segment were not detected.

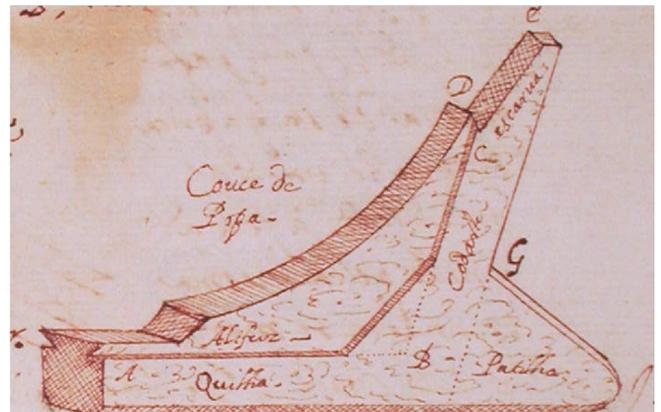

Fig. 4. Image taken from Lavanha's treatise "Livro Primeiro de Arquitectura Naval," c. 1600 depicting a composite object and its parts, including letters referenced in the narrative.

*C. Object Recognition in Images*

Image recognition is a well-established discipline, and with the availability of large collection of images, training such algorithms has produced impressive results. For example, face recognition is used in security systems, driverless cars identify objects on the road, and finding images where a particular object is present is offered by apps in mobile phones.

Once segments in images have been identified, our methodology aims at recognizing the objects in the image and label them. The segmentation step is carried out by the SAM2 algorithm [23, 24, 25, 26, 27, 28]. SAM2 generates boxes with the coordinates where the different objects are located, it also produces more detailed information such as the pixels of the contours of the objects identified. This information is stored in a JSON file, which can be exported for further processing. In step two, we use Grounding DINO, a zero-shot image recognition algorithm that assigns labels to the different objects identified, and also a description of the image (Fig. 7).

*D. Use of Generative AI for Labeling Images*

GenAI methods can be used to generate labels and descriptions of the images. This can be used to augment information about the objects and the entire image. Fig. 7 demonstrates two approaches for labeling and describing segments in treatises. The image corresponds to Bartolomeu Crescenzio's Nautica Mediterranea (1601). The left image shows two boxes labeled "ships boat" and "boat," and also several words highlighted in color in a list. These labels were generated by GroundingDino using the default NLTK and CLIP for label generation.

The image on the right was segmented and labeled using OpenAI's ChatGPT. Notice that the number of boxes identified is greater than the previous method. On the one hand, there is more detail for each box. The list of words is labeled as "terms terms," which is correct, and can be used for further process that area. On the other hand, there are numerous false positives: two pulleys are labeled as "axes," which is not correct. To the far right, the fore edge of the book was labeled as a "sharpened object".

GenAI systems such as chatGPT can be used to associate definitions and explanations of ship parts enhancing the description of images. However, the prompts given to chatGPT have to be carefully crafted. Take for example the definition of "rider frame," which is a component of the ship. When given the prompt "what is a rider frame," the definitions returned belong to three contexts: motorcycles/bycicles, insurance/legal documents, mechanical engineering/suspension systems. The last one is the closest related to ships, which was our focus, yet it is still incomplete and incorrect:

> "In some engineering or robotics contexts, a 'rider frame' might refer to a movable or supportive frame component designed to bear or guide a load (the 'rider') along a path or track, especially in conveyor systems or suspension setups."

We then followed up by providing more context, specifying "in a shipbuilding or nautical context," the response was more detailed including references to its role as structural properties and their purpose and importance for ocean-going vessels.

## V. LEVERAGING ONTOLOGIES FOR IMAGE LABELING

An important observation with the use of prompts for enhancing the description of images and ship components is that if instead of using "random" words as prompts to improve the descriptions, we can leverage words in definitions from a specialized glossary along with concepts and properties from an ontology of nautical architecture. We hypothesize that this approach will provide more focused prompts and hence better results.

An ontology is a formal description of concepts in a domain and properties of those concepts form a knowledge base. Ontologies have been extensively used in many disciplines such as biology, the semantic web, and many more [29, 30, 31, 32, 33, 34, 35]. To this end a group of experts contributed to the development and creation of *ontoShip*, which is a specialized ontology containing nearly 250 concepts about naval architecture of wooden ships, along with numerous categories such as hull components, auxiliary components, joining and fastening systems, to name a few.

In addition to functional properties, this ontology can also be used to identify spatial properties of ship components. For example, if a component is at stern or bow; located at the bottom close to the keel, or on the deck, and based on this information, how components are related to each other. As a way of example, Fig. 6 (found on the next page) depicts more context about a RiderFrame, indicating that it is a Hull Component out of two other categories (Auxiliary Component and Internal Structure) of which it is not part of. RiderFrame is also related to another ship component: Frame.

## VI. RESULTS AND DISCUSSION

Results from our initial experiments demonstrate the potential of image processing techniques for segmenting images. Depending on the algorithm used, results will vary. In our tests, SAM2 generated better segments when compared to Watershed. Some of our earlier approaches actually took care of the tagging and bounding boxes before SAM's segmentation; it also appears that regardless of when the segmentation is done, the results of this step have no impact on the subsequent steps, because the segment masks are stored and then applied to the image, either before or after the tags and boxes are applied to the image. Even using the same model, but changing the settings, different results were generated.

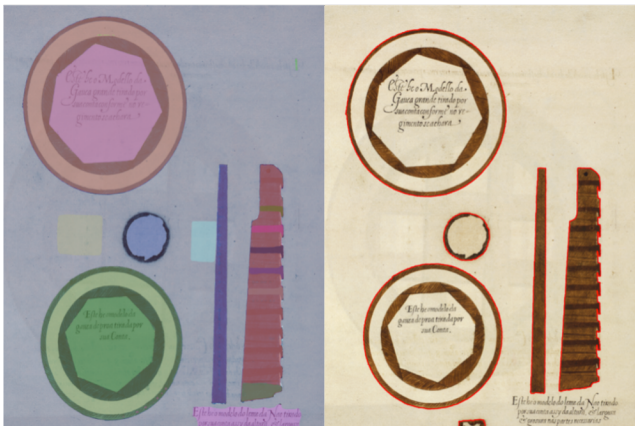

Fig. 5. Comparison of the application of two segmentation algorithms. SAM2 (left) shows geometric shapes in different colors. OpenCV's Watershed (right) shows the contours of several objects detected on the image.

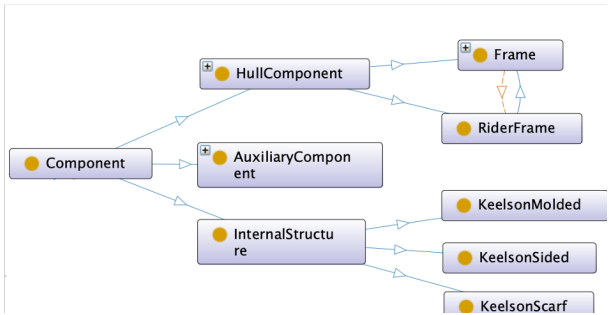

Fig. 6. Partial image of a component of a ship RiderFrame and additional context provided by the Ontology, for example to what part of the ship it belongs (is a Hull Component).

Labeling objects in images proved to be more complex than expected. This can be attributed to the availability of training data in highly specialized domains such as nautical archaeology, which leads to difficulty in recognizing images. In addition, images depicting composite objects are very common in naval architecture, making smaller objects even more difficult to identify.

As far as labeling or tagging images, depending on the algorithm used, the quality of the labels varied substantially.

OpenAI's ChatGPT showed better preliminary results when compared to either the RAM model alone or the combined NLTK and BLIP pipeline initially used. Variation in the quality of labeling results is therefore contingent upon the prompts used and the model's internal functionality. The reason of this variation is due to the training employed in these models, and the prompts used to expand the results, see Fig. 7 for different results (see section IV. C. for more details).

VII. FUTURE WORK

Based on the results obtained, we will process the 14 treatises available in our collection, encompassing approximately 4,000 images in languages such as French, English, Dutch, French, Italian, and Latin, dating from 1550 to 1813. Since Microsoft's Florence-2 is considered a Swiss army knife for CV tasks, such as: generating bounding boxes and using caption-to-phrase grounding to obtain their corresponding labels, we will conduct a comprehensive comparison between Florence-2 with Recognize Anything Model (RAM), BLIP, NLTK, and Tag2Text.

We have observed that the prompts provided to the different labeling algorithms greatly impact the quality of the labels generated. Therefore, we will fully incorporate *glosShip*, a curated multilingual glossary of nautical terms along with concepts and properties expressed in *ontoShip*. We hypothesize that this approach is similar to query expansion, a widely used method to improve results in information retrieval.

We envision to design and evaluate a metric to better quantify and measure the difference between labels and descriptions generated with the different labeling algorithms, compared with the use of *glosShip* and *ontoShip*, which having been edited and curated by human experts could be used as ground truth. Finally, we would like to test different combination of labeling models in our pipeline, for example two preliminary tests we conducted include: 1) using ChatGPT to generate either a one-sentence caption focused on ship parts present in the image or a list of words, then feeding the labels or the one-sentence caption to Florence-2, which then assigns labels to the bounding boxes, and 2) let Florence-2 caption the image by itself, then use caption-to-phrase grounding to generate final labels based on the generated caption. We believe that our segmentation and labeling approach has great potential to improve the curation, indexing, and dissemination of repositories with images of high historical and cultural value.

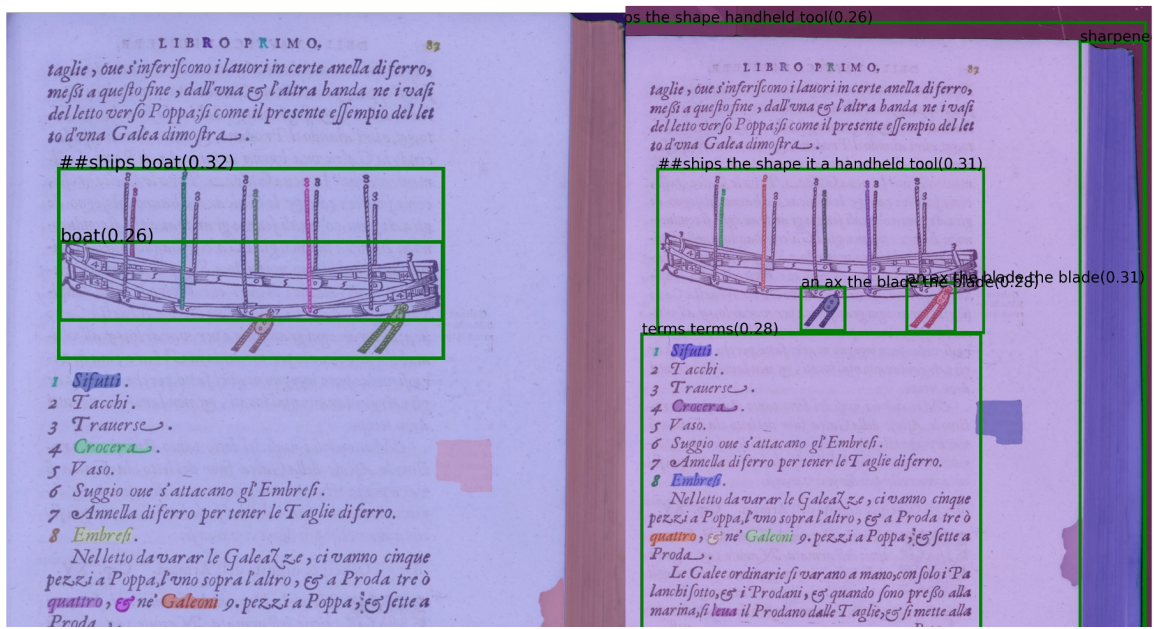

Fig. 7. Results of label generation using two different methods. Results of using default RAM labeling algorithm (left). Labels generated using terms from OpenAI's ChatGPT generated definition (right). Notice that the latter approach identifies more objects (enclosed in the boxes), but at the expense of mislabeled items.

ACKNOWLEDGMENT

We would like to thank Dr. Filipe Castro for providing images of treatises and expertise in nautical archaeology. Dr. Richard Furuta for guidance in data curation and information retrieval. Mr. Dick Steffy for information on the ontology. A group of international maritime archaeologists and numerous scholars and students affiliated with the Center for Maritime Archaeology and Conservation at Texas A&M University worked on the multilingual glossary of nautical terms.